\documentclass{article}
\usepackage{ijcai26}

\usepackage{times}
\usepackage{soul}
\usepackage{url}
\usepackage[hidelinks]{hyperref}
\usepackage[utf8]{inputenc}
\usepackage[small]{caption}
\usepackage{graphicx}
\usepackage{amsmath}
\usepackage{amsthm}
\usepackage{booktabs}
\usepackage{algorithm}
\usepackage{algorithmic}
\usepackage[switch]{lineno}
\usepackage{amssymb}

\newtheorem{definition}{Definition}

\title{DynaOD: Dynamic Origin-Destination Flow Generation with Discrete-to-Continuous Temporal Semantic Modeling}

\author{
Jie Zhao$^1$
\and
Xianqi Dai$^2$\and
Jie Feng$^{3}$\footnote{Corresponding author.}\and
Huandong Wang$^1$
\And
Yong Li$^{1}$\\
\affiliations
$^1$Department of Electronic Engineering, BNRist, Tsinghua University\\
$^2$Tsinghua Shenzhen International Graduate School\\
$^3$Zhongguancun Academy\\
\emails
fengjie@bza.edu.cn, liyong07@tsinghua.edu.cn
}

\begin{document}

\maketitle

\begin{abstract}
Dynamic origin-destination (OD) flow generation seeks to synthesize realistic mobility dynamics from temporal context alone, without relying on historical OD observations. A key challenge is to translate semantic temporal signals into temporally coherent OD patterns while preserving the inherent spatial heterogeneity of urban regions. We propose DynaOD, a semantic-driven framework that models temporal dynamics through two complementary perspectives: discrete directional trends that characterize qualitative shifts in urban activity patterns, and continuous temporal evolution that captures how such shifts unfold over time. By jointly encoding these temporal semantics, the framework constructs time-varying region representations that condition pretrained static OD generators in a lightweight and plug-and-play fashion. This modular design further supports scalable deployment and cross-city transferability. Extensive experiments on large-scale real-world datasets show that our method consistently outperforms representative baselines in both predictive accuracy and distributional fidelity. Code is publicly available at \url{https://github.com/csjiezhao/DynaOD}.
\end{abstract}

\section{Introduction}
\begin{figure}[t]
    \centering
\includegraphics[width=\columnwidth]{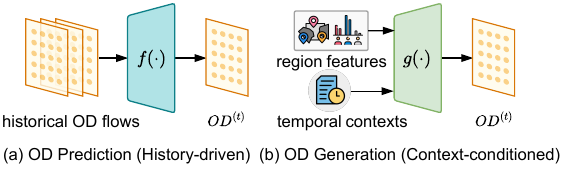}
    \caption{Comparison between history-driven OD prediction and context-conditioned OD generation. OD prediction extrapolates future flows from historical OD flows, whereas OD generation synthesizes flows directly from region attributes and temporal contexts.}
    \label{fig:intro}
\end{figure}

Origin-destination (OD) flows describe collective mobility patterns between geographical regions and constitute a fundamental representation for urban spatial analysis~\cite{andrienko2016revealing}. Accurate OD flows support a wide range of downstream applications, including mobility simulation~\cite{zhang2025noise}, urban planning~\cite{miller2018accessibility}, and policy evaluation~\cite{van2023transport}. When fine-grained trajectory data are unavailable, synthesized OD flows offer an effective alternative for characterizing large-scale urban mobility dynamics~\cite{zhang2020origin,li2021odt}. However, urban mobility exhibits complex temporal dynamics shaped by recurring human routines as well as stochastic external events~\cite{barroso2020correlation}. To capture such variability, dynamic OD flow generation aims to synthesize context-aware OD flows without relying on historical OD observations or real-time traffic measurements. As illustrated in Figure~\ref{fig:intro}, the task is formulated as a conditional generation problem grounded solely in regional semantics and temporal context.

Most existing OD generation methods focus on static OD matrices and primarily model spatial regularities from regional attributes without relying on historical OD observations~\cite{rong2024interdisciplinary}. However, such methods are inherently limited to time-aggregated settings and cannot capture evolving temporal mobility dynamics. Extending OD generation to dynamic temporal scenarios therefore remains largely underexplored~\cite{rong2024learning,yean2025flogan}. Existing efforts typically rely on real-time traffic observations or historical OD sequences~\cite{bauer2017quasi}, which restrict their applicability in cold-start and cross-city settings~\cite{fang2022transfer,li2025cross}. Despite recent progress, dynamic OD generation under minimal supervision remains fundamentally challenging for three reasons.

\textit{First}, temporal context influences OD flows through latent behavioral dynamics rather than explicit flow observations. These effects are often implicit and emerge only after spatial aggregation, making them difficult to capture from limited supervision. \textit{Second}, OD flows exhibit strong spatial heterogeneity that varies across temporal contexts. Regions with similar functional roles may respond similarly under certain contexts yet diverge substantially under others, complicating the context-aware modeling of regional mobility dynamics. \textit{Third}, achieving scalable dynamic OD generation across heterogeneous cities and temporal settings remains difficult. Practical systems must generalize without city-specific retraining while maintaining efficient deployment and low inference overhead.

To address these challenges, we propose \textbf{DynaOD}, a modular framework that models temporal context through semantic temporal abstractions. Specifically, DynaOD disentangles temporal semantics into two complementary components: discrete semantic trends that capture coarse-grained transitions in urban activity patterns, and continuous temporal dynamics that characterize how these transitions evolve over time. By jointly encoding these temporal dynamics, the framework constructs time-varying region features that condition pretrained OD generators for dynamic flow synthesis.
A key advantage of this design is that temporal adaptation can be achieved without modifying the underlying OD generators, enabling seamless integration with diverse pretrained models. In addition, DynaOD incorporates a retrieval-enhanced mechanism to improve cross-city generalization, together with a lightweight distillation strategy for efficient and scalable deployment.

Our contributions are summarized as follows:
\begin{itemize}
    \item We propose a semantic-driven framework that disentangles discrete temporal trends and continuous temporal dynamics for context-aware dynamic OD generation under minimal supervision.
    
    \item We design a retrieval-enhanced regional adaptation mechanism that enables the reuse of regional knowledge across temporal contexts, improving cross-city generalization and generation fidelity.
    
    \item We develop a lightweight plug-in architecture compatible with diverse pretrained OD generators, together with an efficient distillation strategy for scalable deployment.
\end{itemize}

\section{Preliminaries}

\begin{definition}[Region]
An urban area is partitioned into a set of spatial regions $\mathcal{R}=\{r_i\}_{i=1}^{N}$, where each region corresponds to a basic spatial unit such as a census tract or administrative zone.
\end{definition}

\begin{definition}[POI Distribution]
Each region $r_i \in \mathcal{R}$ is associated with a POI distribution vector $\mathbf{p}_i \in \mathbb{R}^{D_p}$ describing the composition of POI categories within the region.
\end{definition}

\begin{definition}[Demographic Attributes]
Each region $r_i \in \mathcal{R}$ is associated with a demographic feature vector $\mathbf{d}_i \in \mathbb{R}^{D_d}$ containing demographic and socio-economic statistics.
\end{definition}

\begin{definition}[Dynamic OD Flow Sequence]
Given a temporal horizon of length $T$, a dynamic OD flow sequence is represented as an ordered set of OD matrices $\{\mathbf{M}^{(t)}\}_{t=1}^{T}$, where $\mathbf{M}^{(t)} \in \mathbb{R}^{N \times N}$ denotes the OD flow matrix at time step $t$.
\end{definition}

\noindent\textbf{Problem Statement.}
Given regions $\mathcal{R}=\{r_i\}_{i=1}^{N}$ with associated POI distributions $\mathbf{p}_i$ and demographic features $\mathbf{d}_i$, together with a temporal context sequence $\mathbf{c}_{1:T}$, the goal of \emph{dynamic OD flow generation} is to synthesize a sequence of OD matrices $\{\mathbf{M}^{(t)}\}_{t=1}^{T}$ that captures realistic urban mobility dynamics under the specified temporal context.

\section{Methodology}
\begin{figure*}[t]
    \centering
    \includegraphics[width=0.85\textwidth]{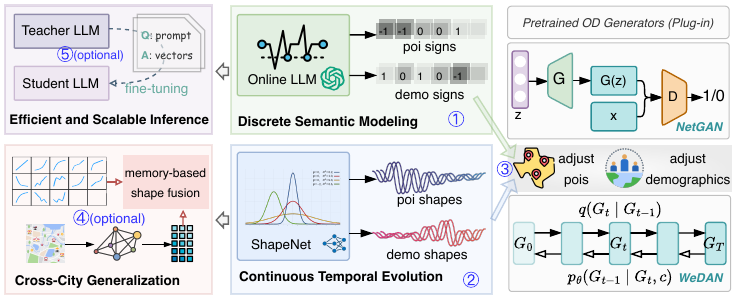}
    \caption{Architecture of DynaOD. (1) An LLM infers discrete directional signals from temporal context;
(2) ShapeNet converts them into continuous multi-day feature evolution;
(3) the resulting features modulate a pretrained OD generator stepwise to produce dynamic OD flows.
Optional components include (4) retrieval-based shape composition for cross-city generalization and
(5) LLM distillation for efficient inference.}
    \label{fig:overview}
\end{figure*}

As illustrated in Figure~\ref{fig:overview}, DynaOD leverages a large language model (LLM) to infer semantic directional trends from temporal context and converts them into continuous temporal feature dynamics via a shape generator.
The resulting features are then used to condition a pretrained static OD generator at each time step, enabling dynamic OD flow synthesis. In addition, the framework supports optional cross-city shape composition and LLM distillation to enhance generalization and deployment efficiency.

\subsection{Discrete Semantic Modeling of Temporal Context}
\label{sec:llm_control}

Dynamic OD generation requires modeling how urban functions and population activities respond to temporal context at a semantic level, rather than directly predicting numerical changes. Temporal cues such as weekdays or holidays often induce qualitative shifts in mobility behaviors~\cite{grinberger2019spatiotemporal,sparks2022shifting}, whose effects on region attributes are indirect and difficult to capture with continuous predictors. Moreover, such directional controls do not have explicit ground-truth supervision and must instead be inferred from contextual semantics and commonsense priors. LLMs are well suited for this task because they encode rich world knowledge and commonsense reasoning capabilities. We therefore introduce a \emph{directional control} mechanism that leverages LLMs to infer coarse-grained semantic trends in urban dynamics.

\paragraph{Discrete directional controls.}
For each region and date, the LLM generates two discrete control vectors: (i) a POI control vector $\mathbf{c}^{\text{poi}} \in \{-1,0,1\}^{D_p}$ and (ii) a demographic control vector $\mathbf{c}^{\text{demo}} \in \{-1,0,1\}^{D_d}$. Each element indicates whether the corresponding feature is expected to increase, remain stable, or decrease under the given spatio-temporal context. Rather than directly conditioning OD flows, these controls encode qualitative directional trends in urban dynamics. The POI control is inferred from regional context and temporal information, while the demographic control is further conditioned on inferred POI trends and demographic attributes, enabling dependency-aware modeling of urban functions and population dynamics.

\paragraph{Design rationale.}
This discrete formulation offers several advantages. It aligns naturally with commonsense reasoning, separates semantic trend inference from continuous magnitude modeling, and produces compact interpretable controls that generalize effectively across cities. Within DynaOD, the inferred directional controls are translated into continuous temporal feature dynamics through a shape generator, enabling pretrained static OD generators to adapt to diverse temporal contexts without retraining.

\subsection{Continuous Temporal Evolution of Region Features}
\label{sec:shapenet}

While directional control specifies \emph{what} should change under a given temporal context, dynamic OD generation further requires modeling \emph{how much} region attributes vary and how such changes evolve over time. We therefore introduce \textbf{ShapeNet}, a differentiable shape generator that translates discrete directional controls into continuous time-varying feature dynamics.

\paragraph{Inputs and outputs.}
For each region $i$ and day $t$ within a window of length $T$, ShapeNet takes as input the static region features $\mathbf{x}_i=[\mathbf{x}_i^{\text{demo}},\mathbf{x}_i^{\text{poi}}]$, directional controls $\mathbf{c}^{\text{poi}}_{i,t}$ and $\mathbf{c}^{\text{demo}}_{i,t}$, and a time-context encoding $\mathbf{u}_{i,t}$. It outputs Gaussian parameters for POI and demographic feature dynamics:
\begin{equation}
(\boldsymbol{\mu}^{\text{poi}}_{i,t}, \log \boldsymbol{\sigma}^{2,\text{poi}}_{i,t}) \in \mathbb{R}^{D_p}, \quad
(\boldsymbol{\mu}^{\text{demo}}_{i,t}, \log \boldsymbol{\sigma}^{2,\text{demo}}_{i,t}) \in \mathbb{R}^{D_d}.
\end{equation}
Shape sequences are sampled through the reparameterization trick and aggregated to obtain continuous temporal shapes $\mathbf{s}^{\text{poi}}_{i,t}$ and $\mathbf{s}^{\text{demo}}_{i,t}$.

\paragraph{Architecture.}
ShapeNet adopts a control- and time-dominant architecture, where static region features are used only for weak conditioning. Specifically, POI controls and demographic controls are first combined with their corresponding activation masks and time-context encodings to form per-day input tokens. These token sequences are independently processed by lightweight temporal encoders to capture day-to-day dependencies, and the resulting hidden representations are subsequently fused into shared temporal embeddings. To incorporate regional heterogeneity without overfitting to city-specific identities, ShapeNet further applies lightweight FiLM~\cite{perez2018film} conditioning based on compact region codes derived from static region features.

\paragraph{Feature modulation.}
Given continuous shapes and discrete controls, DynaOD constructs time-varying region features via multiplicative modulation:
\begin{equation}
\begin{aligned}
    \tilde{\mathbf{x}}^{\text{poi}}_{i,t} &= \mathbf{x}^{\text{poi}}_{i}\odot \left(1 + \mathbf{s}^{\text{poi}}_{i,t}\odot \mathbf{c}^{\text{poi}}_{i,t}\right), \\
    \tilde{\mathbf{x}}^{\text{demo}}_{i,t} &= \mathbf{x}^{\text{demo}}_{i}\odot \left(1 + \mathbf{s}^{\text{demo}}_{i,t}\odot \mathbf{c}^{\text{demo}}_{i,t}\right).
\end{aligned}
\end{equation}
The modulated region features are then fed into a frozen pretrained static OD generator to synthesize dynamic OD flows at each time step.

\paragraph{Training and regularization.}
ShapeNet is trained by minimizing the discrepancy between generated and ground-truth OD flows over the temporal window, while the pretrained OD generator remains frozen. To make optimization tractable, gradients are propagated only through the final DDIM denoising steps during training. We further apply a weighted temporal smoothness regularizer to suppress unrealistic day-to-day variations, with stronger penalties when the corresponding directional controls indicate stability.

\subsection{Cross-City Generalization via Composable Shape Priors}
\label{sec:shapemem}

Although ShapeNet enables continuous temporal feature evolution, directly applying it to unseen cities may suffer from out-of-distribution inference and increased computational overhead. To support cross-city generalization and scalable deployment, we introduce \textbf{ShapeMem}, a retrieval-based memory of composable shape priors that synthesizes region-level shapes without invoking ShapeNet during inference.

\paragraph{ShapeMem construction.}
ShapeNet produces Gaussian temporal shape parameters for POI and demographic features. We aggregate these outputs by weekday to construct region-weekday shape priors, where each memory entry stores Gaussian statistics that capture typical temporal dynamics and their variability. These aggregated priors provide a compact and reusable representation of learned temporal evolution patterns.

\paragraph{Region representation and retrieval.}
To retrieve relevant priors for a target tract, we learn region embeddings using a graph-based representation learning approach~\cite{thakoor2022large}.
During inference, cosine similarity in the embedding space is used to identify top-5 similar tracts from the memory.

\paragraph{Composable prior fusion.}
For a target tract and weekday, retrieved shape priors are fused via similarity-weighted averaging to produce tract-level temporal shapes:
\begin{equation}
\tilde{\mathbf{s}}^{\text{poi}}_{i,w} = \sum_{j\in\mathcal{N}_K(i)} \alpha_{i,j}\,\mathbf{s}^{\text{poi}}_{j,w}, \quad
\tilde{\mathbf{s}}^{\text{demo}}_{i,w} = \sum_{j\in\mathcal{N}_K(i)} \alpha_{i,j}\,\mathbf{s}^{\text{demo}}_{j,w},
\end{equation}
where $\alpha_{i,j}$ denotes normalized similarity weights. This retrieval-based composition serves as a non-parametric approximation of ShapeNet inference for unseen cities.

\paragraph{Plug-and-play inference.}
The composed shapes are injected into the feature modulation step and fed into a pretrained OD generator.
Since ShapeMem operates independently of the OD generator, this mechanism is fully plug-and-play and enables efficient dynamic OD generation across new cities and temporal contexts.

\subsection{Efficient and Scalable Inference}
Dynamic OD generation requires producing directional control vectors for many regions over extended temporal horizons. Although LLMs are effective at inferring such high-level semantic controls, repeatedly querying high-capacity models incurs substantial computational overhead. To improve scalability, we adopt a knowledge distillation strategy that transfers directional control inference from a strong teacher LLM to a lightweight student model for efficient deployment~\cite{cheng2025survey,fang2026knowledge}.

\paragraph{Teacher-student distillation.}
A strong online LLM is first used as a teacher to generate high-quality directional control vectors for POI and demographic attributes.
These outputs are treated as pseudo-labels to supervise a compact student model through supervised fine-tuning. In our implementation, we instantiate the student with Qwen2.5-1.5B-Instruct and fine-tune it using LoRA, updating only the query and value projection matrices. The student is trained on POI and demographic instruction data to reproduce the teacher's discrete directional decisions under identical region-date queries, without requiring explicit semantic descriptions for individual feature dimensions.

\paragraph{Structured directional prediction.}
Although directional control fundamentally relies on semantic understanding and commonsense inference, the distilled student model is trained under a structured discrete prediction formulation for efficient supervised fine-tuning. Specifically, the student predicts fixed-length vectors with entries in $\{-1,0,1\}$ for POI and demographic controls. POI and demographic prediction are treated as two separate instruction formats during training, which simplifies optimization and improves prediction stability while enabling unified deployment.

\paragraph{Plug-and-play deployment.}
After distillation, the student model fully replaces the teacher during inference and serves as a lightweight directional controller. Since the controller is decoupled from both ShapeNet and the OD generator, it can be independently updated or replaced without affecting the remaining components. This modular design enables efficient and scalable dynamic OD generation in large-scale deployment scenarios.

\section{Experiments}
\subsection{Experimental Setup}
\subsubsection{Datasets}
We construct a tract-level dynamic OD flow dataset by integrating publicly available U.S. mobility flows~\cite{kang2020multiscale} with regional attributes from a large-scale commuting benchmark~\cite{rong2025large}. The dataset spans 500 U.S. counties in January 2019, where counties are treated as urban instances and census tracts as spatial regions. Following~\cite{rong2025large}, each tract is associated with POI distributions over 34 categories and 97-dimensional demographic attributes. Daily OD matrices are aligned with tract-level attributes to ensure consistent spatial correspondence.

Each date is associated with an 8-dimensional temporal context vector consisting of a weekday one-hot encoding and a holiday indicator. To evaluate spatial and temporal generalization, we adopt a strict two-axis split protocol. Counties and dates are partitioned into disjoint \emph{seen} and \emph{unseen} sets. Models are trained only on seen counties and seen dates, and evaluated on unseen counties at unseen dates. Under this protocol, only \emph{ShapeNet} is optimized, while the pretrained OD generator and directional control signals remain fixed.

\subsubsection{Metrics}
We evaluate OD generation performance using complementary metrics that capture accuracy, structural consistency, and distributional similarity.
Specifically, \textbf{RMSE} and \textbf{NRMSE} measure element-wise errors, where NRMSE is normalized by the empirical variance of real flows.
\textbf{CPC} (Common Part of Commuting) measures the overlap between predicted and ground-truth OD flows, reflecting structural fidelity.
In addition, \textbf{Jensen Shannon Divergence (JSD)} is computed over inflow, outflow, and full OD flow distributions to assess global mobility pattern similarity.

\subsubsection{Baselines}

We compare DynaOD with a diverse set of representative baselines covering physics-inspired, statistical learning, deep learning, and generative models for OD flow modeling.
The compared methods include:
(1) \emph{Physics-inspired models}: two variants of the gravity model, \textbf{GM-P} and \textbf{GM-E}~\cite{zipf1946p};
(2) \emph{Statistical learning models}: \textbf{Random Forest (RF)}~\cite{pourebrahim2019trip}, \textbf{Support Vector Regression (SVR)}~\cite{rodriguez2021origin}, and \textbf{Gradient Boosting Regression Tree (GBRT)}~\cite{robinson2018machine};
(3) \emph{Deep learning models}: the \textbf{Deep Gravity Model (DGM)}~\cite{simini2021deep};
(4) \emph{Generative graph models}: \textbf{NetGAN}~\cite{bojchevski2018netgan};
and (5) \emph{Graph diffusion-based models}: \textbf{WeDAN}~\cite{rong2025large}.

\paragraph{Adapting baselines to dynamic settings.}
For fair comparison, we augment all methods with the same temporal context features used by DynaOD, concatenated to region-level inputs whenever applicable, without modifying their original architectures. For NetGAN and WeDAN, we follow standard training protocols and use pretrained models~\cite{rong2025large}, denoted as NetGAN$_{pt}$ and WeDAN$_{pt}$.
These models are directly applied at inference time under different temporal contexts without additional fine-tuning.
Unless otherwise stated, DynaOD adopts WeDAN$_{pt}$ as its underlying OD generator.
To demonstrate modularity, we additionally replace the generator with NetGAN$_{pt}$ in subsequent experiments.

\subsection{Performance Comparison}

\begin{table}[t]
\centering
\setlength{\tabcolsep}{1.5pt}
\resizebox{\columnwidth}{!}{
\begin{tabular}{lcccccc}
\toprule
Model & CPC$\uparrow$ & RMSE$\downarrow$ & NRMSE$\downarrow$ & JSD-In$\downarrow$ & JSD-Out$\downarrow$ & JSD-OD$\downarrow$ \\
\midrule
GM-P & 0.181 & 317.11 & 1.161 & 0.722 & 0.834 & 0.662 \\
GM-E & 0.179 & 309.23 & 1.163 & 0.731 & 0.838 & 0.671\\
RF & 0.219 & 304.35 & 1.106 & 0.691 & 0.798 &  0.513\\
SVR & 0.206 & 312.98 & 1.143 & 0.706 & 0.780 & 0.444 \\
GBRT & 0.196 & 310.41 & 1.127 & 0.720 & 0.806 & 0.550\\
DGM & 0.214	& 312.29 & 1.133 & 0.747 & 0.827 & 0.502\\
NetGAN$_{pt}$ & \underline{0.365} & 293.42 & 1.129 & 0.657 & \underline{0.736} & 0.574\\
WeDAN$_{pt}$ & 0.331 & \underline{286.51} & \underline{1.056} & \underline{0.502} & 0.741 & \underline{0.333}\\
DynaOD & \textbf{0.492} & \textbf{253.09} & \textbf{0.940} & \textbf{0.322} & \textbf{0.421}	& \textbf{0.220}\\
\midrule
Improve & 34.7\% & 11.7\% & 11.0\% & 35.9\% & 42.8\% & 33.9\% \\
\bottomrule
\end{tabular}}
\caption{Performance comparison on the test set.
$\uparrow$ ($\downarrow$) indicates that higher (lower) values are better.
\textbf{Bold} and \underline{underline} denote the best and second-best results, respectively.}
\label{tab:basic_results}
\end{table}

Table~\ref{tab:basic_results} reports the performance comparison under the challenging setting of unseen cities and unseen dates. DynaOD consistently outperforms all baselines across every evaluation metric, demonstrating strong robustness to both spatial and temporal distribution shifts. Classical gravity-based and statistical learning methods exhibit limited performance, reflecting the difficulty of modeling dynamic urban mobility under temporally invariant assumptions. Learning-based regressors and DGM provide moderate improvements, yet still struggle to capture evolving temporal mobility patterns. Among generative baselines, pretrained NetGAN and WeDAN remain competitive due to their strong graph-based generation capabilities. Nevertheless, their reliance on static OD distributions limits generalization to dynamic scenarios with unseen spatial-temporal contexts. In contrast, DynaOD improves CPC by \textbf{34.7\%} over the strongest baseline and reduces distributional divergence by up to \textbf{42.8\%}, indicating substantially better alignment with real-world mobility dynamics.
Overall, the results validate the effectiveness of semantic temporal control and continuous temporal adaptation in dynamic OD flow generation.

\subsection{Ablation Study}

\subsubsection{Effectiveness of Directional Control}

\begin{figure}[htbp]
    \centering
\includegraphics[width=\columnwidth]{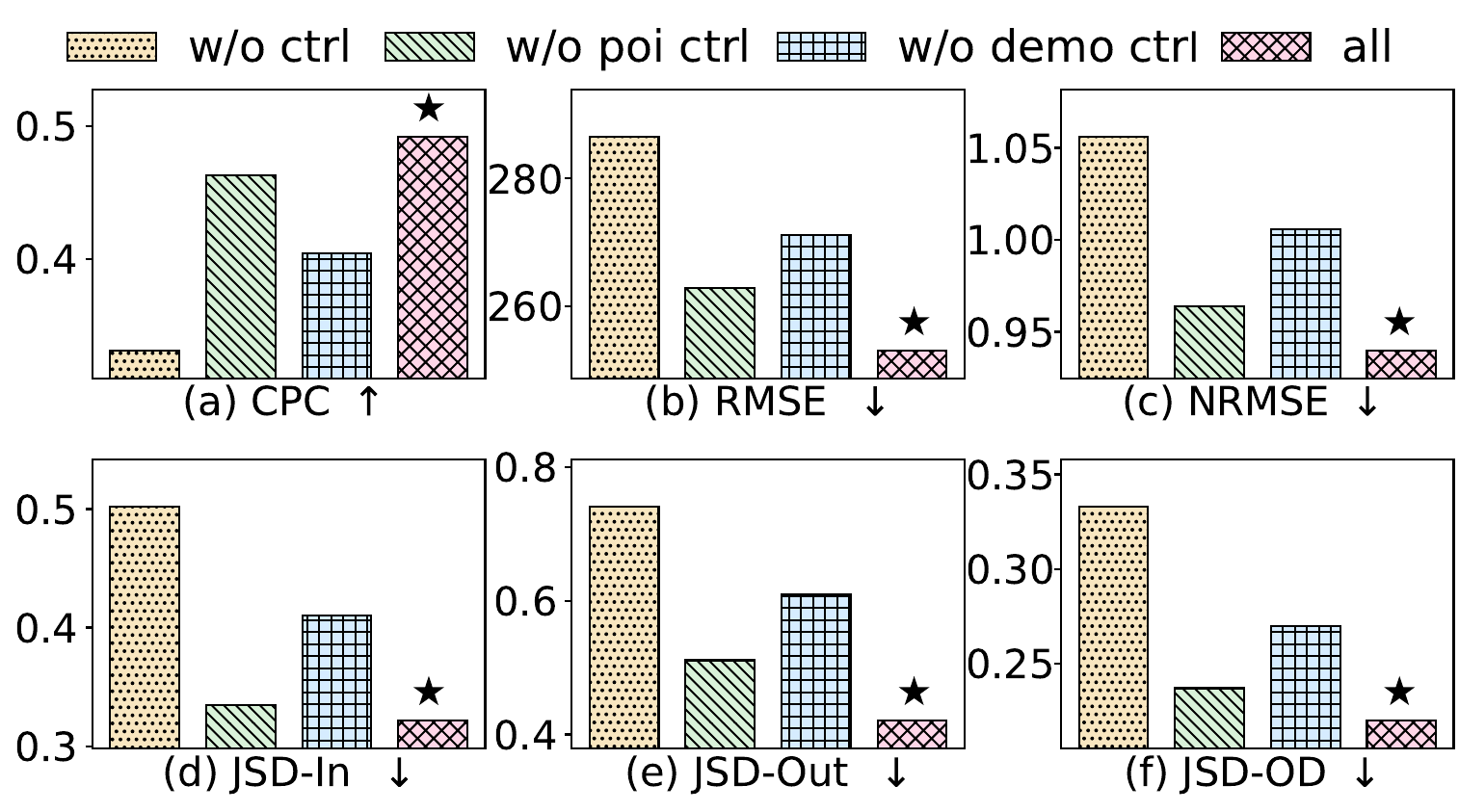}
    \caption{Ablation results on directional control. The star-shaped marker denotes the best performance among all variants.}
    \label{fig:ablation_ctrl}
\end{figure}

Figure~\ref{fig:ablation_ctrl} reports the ablation results on directional control in DynaOD. Removing all directional controls (\emph{w/o ctrl}) leads to substantial degradation across every metric, suggesting that temporal evolution alone is insufficient to model dynamic mobility patterns without semantic guidance.

Introducing either POI control or demographic control consistently improves performance over the no-control variant, demonstrating that both contribute meaningful temporal conditioning signals. POI control mainly enhances reconstruction accuracy, whereas demographic control more effectively reduces distributional discrepancies in inflow and outflow patterns.
When both controls are jointly incorporated, the model achieves the best overall performance across all metrics. The consistent gains over single-control variants indicate that urban functional dynamics and population behavior dynamics provide complementary information for modeling temporal mobility variations.

We further examine whether the effectiveness of directional control stems from semantic reasoning or merely from the discrete control formulation. To this end, we replace the LLM-based controller with a lightweight MLP classifier, resulting in the variant DynaOD$_{\text{CLS}}$. This replacement consistently reduces both reconstruction accuracy and distributional fidelity, with CPC decreasing from 0.492 to 0.413 and JSD-OD increasing from 0.220 to 0.274. These findings indicate that directional control benefits not only from discrete label prediction, but also from the semantic priors provided by LLMs. By contrast, the classifier-based variant relies solely on downstream OD supervision, limiting its ability to generalize across unseen spatial-temporal contexts.

\subsubsection{Effectiveness of Shape Modeling}
\begin{figure}[b]
    \centering
\includegraphics[width=\columnwidth]{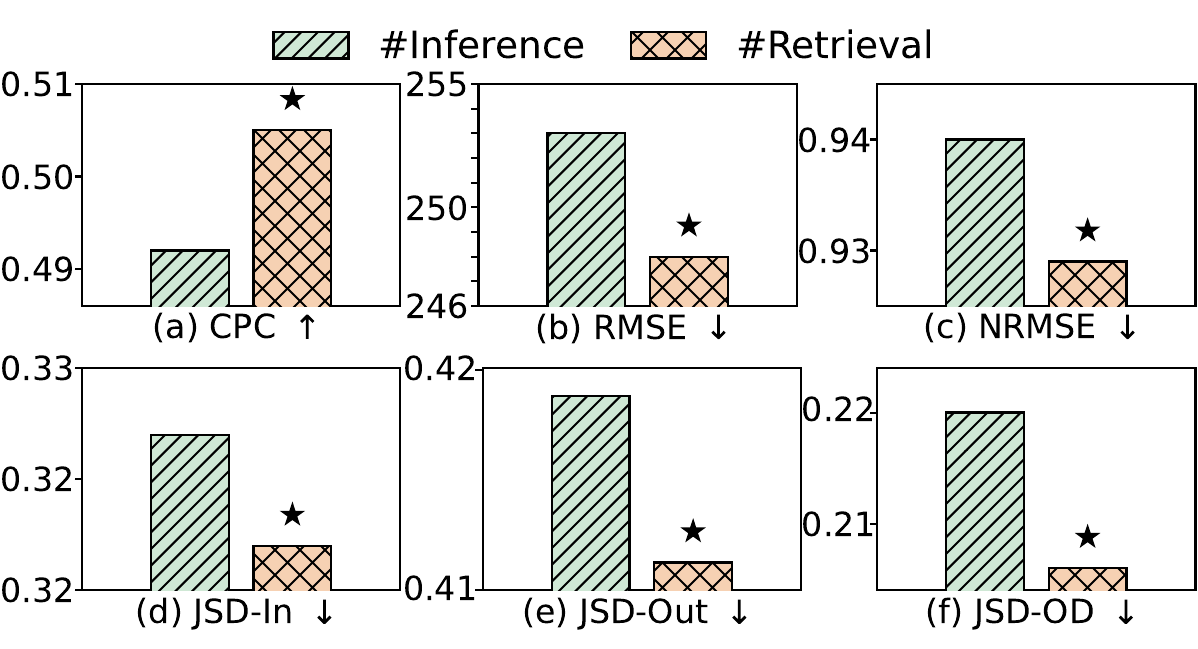}
    \caption{Ablation study on the effectiveness of shape modeling.}
    \label{fig:shape_ablation}
\end{figure}

Figure~\ref{fig:shape_ablation} compares two strategies for continuous shape construction in DynaOD: direct ShapeNet inference (\emph{\#Inference}) and retrieval-based composition using similar regions (\emph{\#Retrieval}). Across all metrics, \emph{\#Retrieval} consistently achieves better performance than \emph{\#Inference}.
The improvements highlight two benefits of retrieval-based shape composition. First, retrieving shapes from similar regions yields more robust temporal dynamics under unseen spatial-temporal settings. Second, \emph{\#Retrieval} removes the need for ShapeNet inference during testing, substantially improving deployment efficiency and scalability. These results demonstrate that composable shape priors provide a scalable solution for cross-city dynamic OD generation.

\subsection{Efficiency and Modularity Analysis}

\subsubsection{LLM Inference Overhead}
We evaluate the computational overhead of LLM-based directional control using gpt-4o-mini. On average, a POI control query requires 1.94s with 415 input tokens and 110 output tokens (approximately \$0.00013), while a demographic control query requires 4.51s with 1436 input tokens and 287 output tokens (approximately \$0.00039). Despite the additional LLM inference cost, directional control generation remains practical because region-level queries are highly parallelizable. Furthermore, the proposed distillation strategy substantially reduces deployment overhead by replacing online LLM inference with a lightweight student model during testing.

\subsubsection{Effectiveness of LLM Distillation}

\begin{table}[t]
\centering

\setlength{\tabcolsep}{1.5pt}
\resizebox{\columnwidth}{!}{
\begin{tabular}{lcccccc}
\toprule
LLM & CPC$\uparrow$ & RMSE$\downarrow$ & NRMSE$\downarrow$ & JSD-In$\downarrow$ & JSD-Out$\downarrow$ & JSD-OD$\downarrow$ \\
\midrule
Qwen-2.5-1.5B & 0.403 & 274.43 & 1.010 & 0.411 & 0.637 & 0.276\\
Qwen-2.5-1.5B-SFT & 0.476 & 260.86 & 0.959 & 0.336 & 0.475 & 0.234\\
GPT-4o-mini & 0.492 & 253.09 & 0.940 & 0.322 & 0.421 & 0.220 \\
\bottomrule
\end{tabular}}
\caption{Performance comparison of DynaOD under different directional control generators (e.g., LLMs).}
\label{tab:llm_dist}
\end{table}

Table~\ref{tab:llm_dist} compares DynaOD under different directional controllers, including GPT-4o-mini, Qwen-2.5-1.5B, and its distilled variant Qwen-2.5-1.5B-SFT. All other components are kept fixed to isolate the effect of directional control quality. GPT-4o-mini achieves the strongest overall performance, whereas the vanilla Qwen-2.5-1.5B shows clear degradation, especially in distributional metrics. This suggests that lightweight models struggle to reliably infer semantic directional signals under unseen spatial-temporal settings. After supervised fine-tuning, Qwen-2.5-1.5B-SFT substantially narrows the gap and achieves performance close to the teacher model.

\begin{figure}[b]
    \centering
\includegraphics[width=0.9\columnwidth]{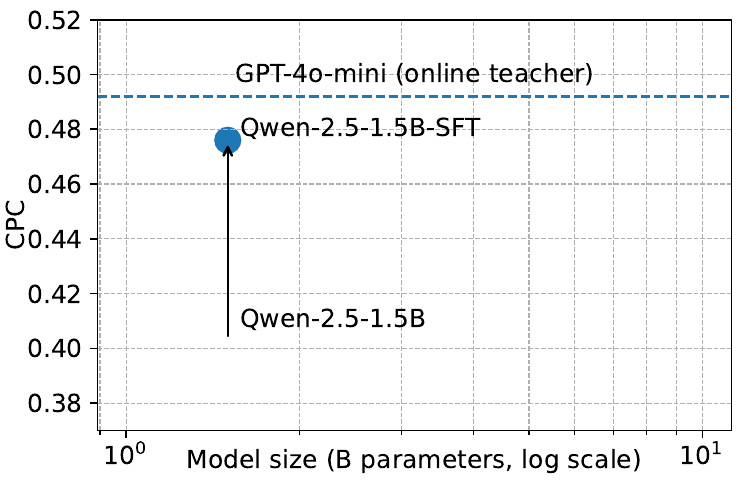}
    \caption{Performance-cost trade-off of directional control models.}
    \label{fig:llm_dist}
\end{figure}

Figure~\ref{fig:llm_dist} further shows the performance-cost trade-off of different controllers. Distillation significantly improves the efficiency-quality trade-off of the lightweight model, enabling efficient local inference while preserving strong generation quality. Overall, the results demonstrate that the proposed distillation strategy effectively transfers semantic control capability from a strong LLM to a student model.

\subsubsection{Generality Across OD Generators}

\begin{table}[t]
\centering
\setlength{\tabcolsep}{1.5pt}
\resizebox{\columnwidth}{!}{
\begin{tabular}{lcccccc}
\toprule
Model & CPC$\uparrow$ & RMSE$\downarrow$ & NRMSE$\downarrow$ & JSD-In$\downarrow$ & JSD-Out$\downarrow$ & JSD-OD$\downarrow$ \\
\midrule
NetGAN$_{pt}$             & 0.365 & 293.42 & 1.129 & 0.657 & 0.736 & 0.574 \\
DynaOD$_{\text{NetGAN}}$  & 0.461 & 281.01 & 1.096 & 0.611 & 0.519 & 0.522 \\
\midrule
WeDAN$_{pt}$              & 0.331 & 286.51 & 1.056 & 0.502 & 0.741 & 0.333 \\
DynaOD$_{\text{WeDAN}}$   & 0.492 & 253.09 & 0.940 & 0.322 & 0.421 & 0.220 \\
\bottomrule
\end{tabular}}
\caption{Generality of DynaOD across different pretrained OD generators.
$\uparrow$ ($\downarrow$) indicates higher (lower) values are better.}
\label{tab:generator_generality}
\end{table}

Table~\ref{tab:generator_generality} evaluates DynaOD across different pretrained OD generators, including NetGAN and WeDAN. In both settings, the generators remain frozen and DynaOD is applied without modifying the original architecture or training procedure.
Across both generators, integrating DynaOD consistently improves performance in terms of CPC, error metrics, and distributional similarity.
For the relatively weaker NetGAN, DynaOD leads to substantial gains, particularly in CPC and JSD metrics, indicating that directional control and shape modulation effectively adapt static graph generators to dynamic mobility scenarios.
For the stronger WeDAN generator, DynaOD further improves performance across all metrics, achieving the best overall results.

Notably, while DynaOD consistently improves different OD generators, the absolute performance remains dependent on the capacity of the underlying generator. This indicates that DynaOD enhances temporal adaptability while preserving the core modeling capability of the base generator. As a result, DynaOD can naturally benefit from future advances in pretrained OD generators.

\begin{figure}[b]
    \centering
\includegraphics[width=\columnwidth]{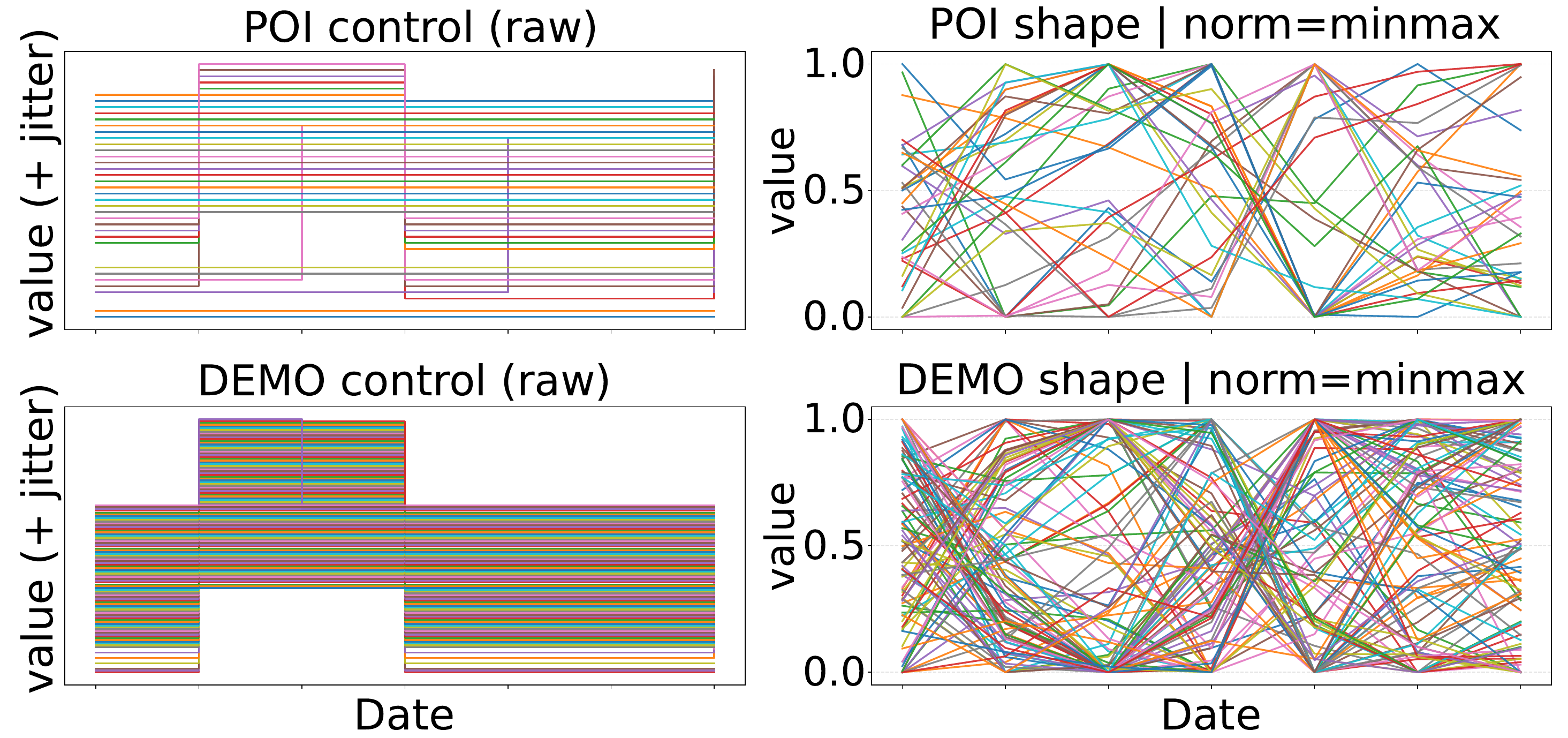}
    \caption{Case study of discrete directional control and continuous shape generation for a randomly selected tract over a 7-day window (2019-01-25 to 2019-01-31).
Left: raw discrete POI and demographic control vectors generated by \texttt{GPT-4o-mini} (with jitter for visualization).
Right: corresponding continuous feature shapes produced by ShapeNet (min-max normalized per dimension).}
\label{fig:case_study}
\end{figure}

\subsection{Case Study}

To qualitatively illustrate how DynaOD translates temporal context into dynamic feature dynamics, we present a case study on a single region over a 7-day generation window. Figure~\ref{fig:case_study} visualizes the discrete directional controls generated by the LLM and the corresponding continuous feature shapes produced by ShapeNet.

The left column shows the discrete POI and demographic control vectors. Although the controls are coarse-grained and temporally piecewise constant, they exhibit stable directional patterns across days. The right column shows the resulting continuous feature trajectories, where smooth temporal evolution naturally emerges from the discrete semantic controls.

This example demonstrates how DynaOD decouples semantic temporal control from continuous numerical modeling: high-level directional controls determine \emph{what} should change, while ShapeNet models \emph{how} these changes evolve smoothly over time. Consequently, DynaOD can generate temporally coherent mobility dynamics without relying on historical OD observations.

\section{Related Work}
Urban spatio-temporal modeling has been widely studied~\cite{li2023dynamic,xu2023continuous,yuan2024unist}, among which OD flow modeling represents an important research direction. Existing studies can be broadly divided into static OD flow generation and dynamic OD flow modeling.

\subsection{Static OD Flow Generation}

Early work on static OD flow generation mainly relied on physical models, such as the gravity model~\cite{zipf1946p} and the radiation model~\cite{simini2012universal,kang2015generalized}, which model population movement through simplified physical assumptions. While computationally efficient, these approaches are constrained by strong inductive assumptions and often fail to capture the complex and heterogeneous nature of human mobility.
Recent studies have shifted toward data-driven OD generation, leveraging machine learning and deep learning models to synthesize OD flows directly from urban attributes such as demographic, socioeconomic, and POI distributions~\cite{liu2020learning,pourebrahim2019trip,robinson2018machine,simini2021deep,rong2023goddag}. These approaches demonstrate that OD flows contain transferable spatial regularities and can be generated without direct trajectory observations. Nevertheless, most existing methods focus on static or time-aggregated settings and are evaluated within individual cities, limiting their ability to generalize across cities and evolving temporal contexts.

\subsection{Dynamic OD Flow Modeling}

Research on dynamic OD flows has primarily focused on forecasting, estimation, and scenario-driven simulation, rather than context-conditioned generation. Limited studies consider scenario-based OD generation. For instance, \cite{yean2025flogan} uses conditional GANs to generate mobility flows under simulated urban scenarios with adaptive region definitions. However, such approaches still rely on historical mobility data for calibration and cannot generalize to unseen temporal contexts.

Most existing methods follow the forecasting paradigm, inferring future OD matrices from historical OD sequences using spatio-temporal learning models, including dynamic graph learning~\cite{zhang2021dynamic,zhang2022dynamic}, continuous- and discrete-time representation learning~\cite{xu2023continuous}, hypergraph-based modeling~\cite{shen2024short}, and prototype-based hierarchical learning~\cite{yuan2025spatio}. While effective for short-term forecasting, these approaches fundamentally depend on sufficient historical OD observations.
Another line of research estimates or imputes dynamic OD flows from indirect traffic observations. Representative approaches integrate traffic states into OD modeling~\cite{rong2024learning}, transfer knowledge across heterogeneous data sources~\cite{chen2024dynamic}, or formulate OD estimation as traffic-constrained optimization problems~\cite{englezou2024path}. Despite reducing reliance on direct OD observations, these methods still require partial traffic measurements or calibrated physical assumptions.

Overall, existing dynamic OD modeling methods remain heavily dependent on historical OD data, traffic observations, or scenario-specific calibration. In contrast, dynamic OD generation under minimal supervision, where temporal context alone drives OD generation, remains largely under-explored.

\section{Limitations}
Despite its effectiveness, DynaOD has several limitations.
First, the quality of discrete control depends on the reasoning capability of the language model, which may vary across regions with sparse or ambiguous contextual information.
Second, ShapeNet currently models temporal evolution at a fixed daily granularity, and extending it to finer or irregular time scales remains an open problem.
Finally, while DynaOD is generator-agnostic, its absolute performance is still bounded by the expressiveness of the underlying OD generator.

\section{Conclusion}

This paper introduces DynaOD, a framework for dynamic origin-destination flow generation that unifies semantic temporal control with generative mobility modeling. DynaOD uses LLMs to infer interpretable directional controls from temporal context and converts them into continuous feature dynamics through ShapeNet. These time-varying features are then used to adapt pretrained static OD generators for dynamic mobility synthesis without retraining. To improve scalability and cross-city generalization, DynaOD further integrates retrieval-based shape priors and a lightweight LLM distillation strategy. Extensive experiments on large-scale real-world datasets demonstrate that DynaOD consistently outperforms representative baselines under unseen spatial-temporal settings and remains compatible with different pretrained OD generators.

\section*{Acknowledgements}
This work was supported in part by the National Key Research and Development Program of China (Grant No. 2024YFC3307603), and in part by the Science and Technology Innovation Program of Xiongan New Area (Grant No. 2025XAGG0041).

%% The file named.bst is a bibliography style file for BibTeX 0.99c
\bibliographystyle{named}
\bibliography{ijcai26}

@article{kang2020multiscale,
  title     = {Multiscale Dynamic Human Mobility Flow Dataset in the U.S. during the COVID-19 Epidemic},
  author    = {Kang, Yuhao and Gao, Song and Liang, Yunlei and Li, Mingxiao and Kruse, Jake},
  journal   = {Scientific Data},
  volumn    = {7},
  issue     = {390},
  pages     = {1--13},
  year = {2020}
}

@inproceedings{rong2025large,
  title={A Large-scale Dataset and Benchmark for Commuting Origin-Destination Flow Generation},
  author={Rong, Can and Ding, Jingtao and Liu, Yan and Li, Yong},
  booktitle={The Thirteenth International Conference on Learning Representations},
  pages = {96180--96205},
  volume = {2025},
  year={2025}
}

@inproceedings{bojchevski2018netgan,
  title={{NetGAN}: Generating graphs via random walks},
  author={Bojchevski, Aleksandar and Shchur, Oleksandr and Z{\"u}gner, Daniel and G{\"u}nnemann, Stephan},
  booktitle={International conference on machine learning},
  pages={610--619},
  year={2018},
  organization={PMLR}
}

@article{zipf1946p,
  title={The P 1 P 2/D hypothesis: on the intercity movement of persons},
  author={Zipf, George Kingsley},
  journal={American sociological review},
  volume={11},
  number={6},
  pages={677--686},
  year={1946},
  publisher={JSTOR}
}

@article{pourebrahim2019trip,
  title={Trip distribution modeling with Twitter data},
  author={Pourebrahim, Nastaran and Sultana, Selima and Niakanlahiji, Amirreza and Thill, Jean-Claude},
  journal={Computers, Environment and Urban Systems},
  volume={77},
  pages={101354},
  year={2019},
  publisher={Elsevier}
}

@inproceedings{robinson2018machine,
  title={A machine learning approach to modeling human migration},
  author={Robinson, Caleb and Dilkina, Bistra},
  booktitle={Proceedings of the 1st ACM SIGCAS Conference on Computing and Sustainable Societies},
  pages={1--8},
  year={2018}
}

@article{simini2021deep,
  title={A deep gravity model for mobility flows generation},
  author={Simini, Filippo and Barlacchi, Gianni and Luca, Massimilano and Pappalardo, Luca},
  journal={Nature communications},
  volume={12},
  number={1},
  pages={6576},
  year={2021},
  publisher={Nature Publishing Group UK London}
}

@article{rodriguez2021origin,
  title={Origin--destination matrix estimation and prediction from socioeconomic variables using automatic feature selection procedure-based machine learning model},
  author={Rodr{\'\i}guez-Rueda, PJ and Ruiz-Aguilar, JJ and Gonz{\'a}lez-Enrique, J and Turias, I},
  journal={Journal of urban planning and development},
  volume={147},
  number={4},
  pages={04021056},
  year={2021},
  publisher={American Society of Civil Engineers}
}

@article{rong2024learning,
title={Learning to generate temporal origin-destination flow based-on urban regional features and traffic information},
author={Rong, Can and Liu, Zhicheng and Ding, Jingtao and Li, Yong},
journal={ACM Transactions on Knowledge Discovery from Data},
volume={18},
number={6},
pages={1--17},
year={2024},
publisher={ACM New York, NY}
}

@article{simini2012universal,
  title={A universal model for mobility and migration patterns},
  author={Simini, Filippo and Gonz{\'a}lez, Marta C and Maritan, Amos and Barab{\'a}si, Albert-L{\'a}szl{\'o}},
  journal={Nature},
  volume={484},
  number={7392},
  pages={96--100},
  year={2012},
  publisher={Nature Publishing Group UK London}
}

@inproceedings{liu2020learning,
  title={Learning geo-contextual embeddings for commuting flow prediction},
  author={Liu, Zhicheng and Miranda, Fabio and Xiong, Weiting and Yang, Junyan and Wang, Qiao and Silva, Claudio},
  booktitle={Proceedings of the AAAI conference on artificial intelligence},
  volume={34},
  pages={808--816},
  year={2020}
}

@article{rong2023goddag,
  title={{GODDAG}: Generating origin-destination flow for new cities via domain adversarial training},
  author={Rong, Can and Feng, Jie and Ding, Jingtao},
  journal={IEEE Transactions on Knowledge and Data Engineering},
  volume={35},
  number={10},
  pages={10048--10057},
  year={2023},
  publisher={IEEE}
}

@article{yean2025flogan,
  title={{FloGAN}: Scenario-Based Urban Mobility Flow Generation via Conditional GANs and Dynamic Region Decoupling},
  author={Yean, Seanglidet and Zhou, Jiazu and Lee, Bu-Sung and Schl{\"a}pfer, Markus},
  journal={arXiv preprint arXiv:2507.12053},
  year={2025}
}

@article{chen2024dynamic,
  title={Dynamic origin-destination flow imputation using feature-based transfer learning},
  author={Chen, Peng and Wang, Ziyan and Zhou, Bin and Yu, Guizhen},
  journal={IEEE Transactions on Intelligent Transportation Systems},
  year={2024},
  volume={25},
  number={11},
  pages={17147--17159},
  publisher={IEEE}
}

@article{englezou2024path,
  title={Path-based origin-destination matrix estimation utilizing macroscopic traffic dynamics},
  author={Englezou, Yiolanda and Timotheou, Stelios and Panayiotou, Christos G},
  journal={IEEE Transactions on Intelligent Transportation Systems},
  volume={25},
  number={8},
  pages={8819--8836},
  year={2024},
  publisher={IEEE}
}

@article{xu2023continuous,
  title={Continuous-time and discrete-time representation learning for origin-destination demand prediction},
  author={Xu, Yi and Han, Liangzhe and Zhu, Tongyu and Sun, Leilei and Du, Bowen and Lv, Weifeng},
  journal={IEEE Transactions on Intelligent Transportation Systems},
  volume={25},
  number={3},
  pages={2382--2393},
  year={2023},
  publisher={IEEE}
}

@article{shen2024short,
  title={Short-term metro origin-destination passenger flow prediction via spatio-temporal dynamic attentive multi-hypergraph network},
  author={Shen, Loutao and Li, Junyi and Chen, Yong and Li, Chuanjia and Chen, Xiqun and Lee, Der-Horng},
  journal={IEEE Transactions on Intelligent Transportation Systems},
  volume={25},
  number={8},
  pages={9945--9957},
  year={2024},
  publisher={IEEE}
}

@inproceedings{zhang2022dynamic,
  title={Dynamic graph learning based on hierarchical memory for origin-destination demand prediction},
  author={Zhang, Ruixing and Han, Liangzhe and Liu, Boyi and Zeng, Jiayuan and Sun, Leilei},
  booktitle = {Proceedings of the Thirty-First International Joint Conference on Artificial Intelligence, {IJCAI-22}},
  pages     = {2383--2389},
  year      = {2022}
}

@inproceedings{yuan2025spatio,
  title={Spatio-temporal prototype-based hierarchical learning for OD demand prediction},
  author={Yuan, Shilu and Li, Xiaoyu and Mu, Wenqian and Zhong, Ji and Chen, Meng and Sun, Haoliang and Gong, Yongshun},
  booktitle={Proceedings of the Thirty-Fourth International Joint Conference on Artificial Intelligence},
  pages={3597--3605},
  year={2025}
}

@article{andrienko2016revealing,
  title={Revealing patterns and trends of mass mobility through spatial and temporal abstraction of origin-destination movement data},
  author={Andrienko, Gennady and Andrienko, Natalia and Fuchs, Georg and Wood, Jo},
  journal={IEEE transactions on visualization and computer graphics},
  volume={23},
  number={9},
  pages={2120--2136},
  year={2016},
  publisher={IEEE}
}

@inproceedings{zhang2025noise,
  title={{Noise Matters}: Diffusion Model-based Urban Mobility Generation with Collaborative Noise Priors},
  author={Zhang, Yuheng and Yuan, Yuan and Ding, Jingtao and Yuan, Jian and Li, Yong},
  booktitle={Proceedings of the ACM on Web Conference 2025},
  pages={5352--5363},
  year={2025}
}

@book{van2023transport,
  title={The transport system and transport policy: an introduction},
  author={van Wee, Bert and Annema, Jan A and Banister, David and Pud{\=a}ne, Baiba},
  year={2023},
  publisher={Edward Elgar Publishing}
}

@article{li2021odt,
  title={{ODT FLOW}: Extracting, analyzing, and sharing multi-source multi-scale human mobility},
  author={Li, Zhenlong and Huang, Xiao and Hu, Tao and Ning, Huan and Ye, Xinyue and Huang, Binghu and Li, Xiaoming},
  journal={Plos one},
  volume={16},
  number={8},
  pages={e0255259},
  year={2021},
  publisher={Public Library of Science San Francisco, CA USA}
}

@article{barroso2020correlation,
  title={Correlation analysis of day-to-day origin-destination flows and traffic volumes in urban networks},
  author={Barroso, Joana Maia Fernandes and Albuquerque-Oliveira, Jo{\~a}o Lucas and Oliveira-Neto, Francisco Moraes},
  journal={Journal of Transport Geography},
  volume={89},
  pages={102899},
  year={2020},
  publisher={Elsevier}
}

@article{rong2024interdisciplinary,
  title={An interdisciplinary survey on origin-destination flows modeling: Theory and techniques},
  author={Rong, Can and Ding, Jingtao and Li, Yong},
  journal={ACM Computing Surveys},
  volume={57},
  number={1},
  pages={1--49},
  year={2024},
  publisher={ACM New York, NY}
}

@article{grinberger2019spatiotemporal,
  title={Spatiotemporal contingencies in tourists’ intradiurnal mobility patterns},
  author={Grinberger, A Yair and Shoval, Noam},
  journal={Journal of Travel Research},
  volume={58},
  number={3},
  pages={512--530},
  year={2019},
  publisher={SAGE Publications Sage CA: Los Angeles, CA}
}

@inproceedings{perez2018film,
  title={Film: Visual reasoning with a general conditioning layer},
  author={Perez, Ethan and Strub, Florian and De Vries, Harm and Dumoulin, Vincent and Courville, Aaron},
  booktitle={Proceedings of the AAAI conference on artificial intelligence},
  volume={32},
  year={2018}
}

@article{thakoor2022large,
 title={Large-scale representation learning on graphs via bootstrapping},
 author={Thakoor, Shantanu and Tallec, Corentin and Azar, Mohammad Gheshlaghi and Azabou, Mehdi and Dyer, Eva L and Munos, Remi and Veli{\v{c}}kovi{\'c}, Petar and Valko, Michal},
 journal={International Conference on Learning Representations (ICLR)},
 year={2022}
}

@article{fang2026knowledge,
  title={Knowledge distillation and dataset distillation of large language models: Emerging trends, challenges, and future directions},
  author={Fang, Luyang and Yu, Xiaowei and Cai, Jiazhang and Chen, Yongkai and Wu, Shushan and Liu, Zhengliang and Yang, Zhenyuan and Lu, Haoran and Gong, Xilin and Liu, Yufang and others},
  journal={Artificial Intelligence Review},
  volume={59},
  number={1},
  pages={17},
  year={2026},
  publisher={Springer}
}

@article{bauer2017quasi,
  title={Quasi-dynamic estimation of OD flows from traffic counts without prior OD matrix},
  author={Bauer, Dietmar and Richter, Gerald and Asamer, Johannes and Heilmann, Bernhard and Lenz, Gernot and K{\"o}lbl, Robert},
  journal={IEEE Transactions on Intelligent Transportation Systems},
  volume={19},
  number={6},
  pages={2025--2034},
  year={2017},
  publisher={IEEE}
}

@article{zhang2020origin,
  title={Origin-Destination-Based Travel Time Reliability under Different Rainfall Intensities: An Investigation Using Open-Source Data},
  author={Zhang, Qi and Chen, Hong and Liu, Hongchao and Li, Wei and Zhang, Yibin},
  journal={Journal of Advanced Transportation},
  volume={2020},
  number={1},
  pages={8816020},
  year={2020},
  publisher={Wiley Online Library}
}

@article{miller2018accessibility,
  title={Accessibility: measurement and application in transportation planning},
  author={Miller, Eric J},
  journal={Transport Reviews},
  volume={38},
  number={5},
  pages={551--555},
  year={2018},
  publisher={Taylor \& Francis}
}

@article{sparks2022shifting,
  title={Shifting temporal dynamics of human mobility in the United States},
  author={Sparks, Kevin and Moehl, Jessica and Weber, Eric and Brelsford, Christa and Rose, Amy},
  journal={Journal of transport geography},
  volume={99},
  pages={103295},
  year={2022},
  publisher={Elsevier}
}

@article{kang2015generalized,
  title={A generalized radiation model for human mobility: spatial scale, searching direction and trip constraint},
  author={Kang, Chaogui and Liu, Yu and Guo, Diansheng and Qin, Kun},
  journal={PloS one},
  volume={10},
  number={11},
  pages={e0143500},
  year={2015},
  publisher={Public Library of Science San Francisco, CA USA}
}

@article{cheng2025survey,
  title={Survey on Efficient Large Language Models: Principles, Algorithms, Applications, and Open Issues},
  author={Cheng, Jian and Kang, Haidong and Shao, Yuxin and Li, Nan and Chen, Pengjun and Wang, Rui and Long, Saiqin and Yang, Xiaochun and Ma, Lianbo},
  journal={IEEE Transactions on Neural Networks and Learning Systems},
  year={2025},
  publisher={IEEE}
}

@article{zhang2021dynamic,
  title={Dynamic auto-structuring graph neural network: A joint learning framework for origin-destination demand prediction},
  author={Zhang, Dapeng and Xiao, Feng},
  journal={IEEE Transactions on Knowledge and Data Engineering},
  volume={35},
  number={4},
  pages={3699--3711},
  year={2021},
  publisher={IEEE}
}

@inproceedings{li2025cross,
  title={Cross City Traffic Flow Generation via Retrieval Augmented Diffusion Model},
  author={Li, Yudong and Wang, Jingyuan and Yu, Xie and Wang, Peiyu and Huang, Qian},
  booktitle={The Thirty-ninth Annual Conference on Neural Information Processing Systems},
  year={2025}
}

@inproceedings{fang2022transfer,
  title={When Transfer Learning Meets Cross-City Urban Flow Prediction: Spatio-Temporal Adaptation Matters.},
  author={Fang, Ziquan and Wu, Dongen and Pan, Lu and Chen, Lu and Gao, Yunjun},
  booktitle={IJCAI},
  volume={22},
  pages={2030--2036},
  year={2022}
}

@article{li2023dynamic,
  title={Dynamic graph convolutional recurrent network for traffic prediction: Benchmark and solution},
  author={Li, Fuxian and Feng, Jie and Yan, Huan and Jin, Guangyin and Yang, Fan and Sun, Funing and Jin, Depeng and Li, Yong},
  journal={ACM Transactions on Knowledge Discovery from Data},
  volume={17},
  number={1},
  pages={1--21},
  year={2023},
  publisher={ACM New York, NY}
}

@inproceedings{yuan2024unist,
  title={Unist: A prompt-empowered universal model for urban spatio-temporal prediction},
  author={Yuan, Yuan and Ding, Jingtao and Feng, Jie and Jin, Depeng and Li, Yong},
  booktitle={Proceedings of the 30th ACM SIGKDD conference on knowledge discovery and data mining},
  pages={4095--4106},
  year={2024}
}

\end{document}